# MCA Learning Algorithm for Incident Signals Estimation: A Review

Rashid Ahmed, John A. Avaritsiotis
School of Electrical and Computer Engineering, National Technical University of Athens,
9, Iroon Polytechniou St., 15773 Athens, Greece
e-mail: rashidis@mail.ntua.gr

*Abstract*- **Recently there has been many works on adaptive subspace filtering in the signal processing literature. Most of them are concerned with tracking the signal subspace spanned by the eigenvectors corresponding to the eigenvalues of the covariance matrix of the signal plus noise data. Minor Component Analysis (MCA) is important tool and has a wide application in telecommunications, antenna array processing, statistical parametric estimation, etc. As an important feature extraction technique, MCA is a statistical method of extracting the eigenvector associated with the smallest eigenvalue of the covariance matrix. In this paper, we will present a MCA learning algorithm to extract minor component from input signals, and the learning rate parameter is also presented, which ensures fast convergence of the algorithm, because it has direct effect on the convergence of the weight vector and the error level is affected by this value. MCA is performed to determine the estimated DOA. Simulation results will be furnished to illustrate the theoretical results achieved.**

*Index Terms*-**Direction of Arrival; Neural networks; Principle Component Analysis; Minor Component Analysis.**

## I. INTRODUCTION

Neural networks (NNs) have been applied to a wide variety of real–world problems in many applications. The attractive and flexible characteristics of (NNs), such as their parallel operation, learning by example, associative memory, multi-factorial optimization and extensibility, make them well suited to the analysis of biological and medical signals [1,2,3]. A neural network is an information–processing system that has certain performance characteristics in common with biological neural networks. Many methods for the estimation of the Direction of Arrival (DOA) have been proposed including the Maximum Likelihood (ML) technique [4], the minimum variance method of capon [5], the minimum norm method of Reddy [6],Multiple Signal Classification (MUSIC),[7], Estimation of Signal Parameters Via Rotational Invariant Techniques (ESPRIT)[8,9]. The minor component is the direction in which the data have the smallest variance. Although eigenvalue decomposition or singular value decomposition can be used to extract minor component, these traditional matrix algebraic approaches are usually unsuitable for high-dimensional online input data. Neural networks can be used to solve the task of MCA learning algorithm. Other classical methods involve costly matrix inversions, as well as poor estimation performance when the signal to noise ratio and number of samples are small and too large, respectively [10].

In many practical applications, a PCA algorithm deteriorates with decreasing signal to noise ratio[11]. For this reason, we need to handle this situation in order to overcome the divergence problem. In this context, we present a MCA learning algorithm that has a low computational complexity and allows extracting the smallest eigenvalue and eigenvector from input signals, which can be used to estimate DOA.

The paper is organized as follows. In Section II, we discuss the array signal model, and we also describe a theoretical review of some existing Principal Component Analysis (PCA) and Minor Component Analysis (MCA) algorithms. In Section III, we present the model for the DOA measurements. Simulations of results are included in Section IV to evaluate the convergence of the algorithms and some simulation results are presented to illustrate the theoretical results achieved. Finally, conclusions are drawn in Section V.

## II. SIGNAL MODEL and LEARNING ALGORITHMS FOR PCA AND MCA

### A. Signal Model

Consider an array of omnidirectional sensors. The medium is assumed to be isotropic and non-dispersive. Since far-field source targets are assumed, the source wave front scan be approximated by plane waves. Then, for narrow band source signals, we can express the sensor outputs as the sum of the shifted versions of the source signals.
Consider a Uniform Linear Array (ULA) of ($m$) omnidirectional sensors illuminated by $l$ narrow-band signals ($l<m$). At the $l$'th snapshot the output of the $i$'th sensor may be described by [12]

$$X = \sum_{i=1}^{d} \cos 2\pi l d f_i \exp(\sqrt{-1} * (i-1) * 2\pi \Delta \sin(\pi - \theta_i)) \quad (1)$$

Where $\Delta$ is the space between two adjacent sensors, $\theta_i$ the angle of arrival, $d$ signals incident onto the array, $df_i$ normalizes frequency. The incoming waves are assumed to be planned. The output of array sensors is affected by white noise which is assumed to be uncorrelated with the incoming signals. In vector notation, the output of the array results from $l$ complex signals can be written as:

$$x(n) = c(\theta)s(n) + N(n)$$

Where the vectors
$s(n)$: signal vector , $N(n)$: a noise vectorare defined as:

$$x(n) = [x_1(n), \ldots\ldots\ldots\ldots\ldots, x_m(n)]^T_{m\times 1}$$
$$s(n) = [s_1(n), \ldots\ldots\ldots\ldots\ldots, s_l(n)]^T_{l\times 1}$$
$$N(n) = [N_1(n), \ldots\ldots\ldots\ldots\ldots, N_m(n)]^T_{m\times 1}$$

And $C(\theta)$ is the matrix of steering vectors,

is the target *DOA* parameter vector,

$$C(\theta_i) = [C(\theta_1), \ldots\ldots\ldots\ldots\ldots, C(\theta_i)]_{m\times l}$$

Moreover,

$$C(\theta) = exp[-j2\pi i sin\theta/v] \quad (2)$$
$$v = speedlight$$

### B. Learning Algorithm for PCA

Consider the linear neural unit described by
$$y(t) = w^T. X(t) \text{ where } X \in R$$
Where the input vector, $w \in IR$ represents the weight vectors and *y* denotes the neuron's output. The unit is used for extracting the first principal component from the input random signal, that is $y(t)$ should represent $X(t)$ in the best way, in the sense that the expectation error should be minimized.

$$E_x\left[\frac{\|x-yw\|^2}{w}\right]$$

Here $E_x[./w]$ denotes mathematical expectation with respect to $x$ under the hypothesis $w$. The problem may be expressed as,

Solve: $\min E_x[\|x\|^2 - E_x\left[\frac{y^2}{w}\right]$ under $ww^T = 1$ (4)

Consider the feed forward network shown in Fig.1. The following two assumptions of a structural network are made:
- Each neuron in the output layer of the network is linear.
- The network has *m* inputs and *l* output, both of which are specified. Moreover the network has fewer outputs than inputs (i.e. *l<m*).

The only aspect of the network that is subject to training is the set of synaptic weights $w_{ji}$ connecting source nodes *i*, in the input layer to computation nodes *j* in the output layer, where $i = 0,1,\ldots,m$ and $j = 0,1,\ldots l$.

The output $y_j(n)$ of neuron j at time, produced in response to the set of inputs $\{x_i(n)\}_{i=1}^m$, is given by

$$y_j(n) = \sum_{i=1}^m w_{ji}(n) x_i(n) \quad (5)$$

The synaptic weight $w_{ji}$ is adapted in accordance with a generalized form of Hebbian learning [13] according to PCA as shown by:

$$\Delta w_{ji}(n) = \eta \left[y_j(n)x_i(n) - y_j(n) \sum_{k=1}^j w_{ki}(n) y_k(n)\right] \quad (6)$$

Where $\Delta w_{ji}(n)$, is the change applied to the synaptic weight $w_{ji}(n)$ at time, and $\eta$ is the learning rate parameter, greater than zero.

This principal component analysis algorithm has been found very useful for extracting the most representative low-dimensional subspace from a high–dimensional vector space. It is widely employed to analyze multidimensional input vector of hundreds of different stock prices, however when used in signal processing this algorithm deteriorates with decreasing signal to noise ratio[11].

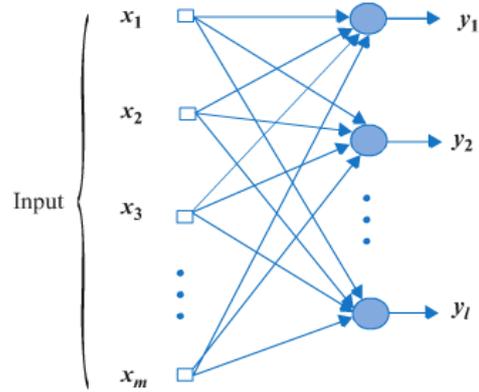

Figure.1: Oja's single-layer linear neural network.

### C. Learning Algorithm for MCA

The opposite of PCA is Minor Component Analysis (MCA), is a statistical method of extracting the eigenvector associated with the smallest eigenvalue of the covariance matrix of input signals. As an important tool for signal processing and data analysis, MCA has been widely applied to: total least squares (TLS) [14], clutter cancellation [15], curve and surface fitting [16], digital beamforming [17], bearing estimation [18], etc. One single linear neuron can be used to extract minor component from input signals adaptively and the eigenvector associated with the smallest eigenvalue of the covariance matrix is called Minor Component, where one seeks to find these directions that minimize the projection variance. These directions are the eigendirections corresponding to the minimum eigenvalue. The applications of MCA arise in total least square and eigenvalue-based spectral estimation methods [19,20]. It allows the extraction of the first minor component from a stationary multivariate random process based on the definition of cost function to be minimized under right constraints. The extraction of the least principal component is usually referred to as MCA. For first Minor Component, what must be found is the weight vector that minimizes the power $E_x\left[\frac{y^2}{W}\right]$ of neurons output.

For convenience, we produce a cost function for minor component estimation, that the problem is minimizing the cost function

$$\min_w \left\{J(w) = \frac{1}{2} E_x[y^2/w + \lambda/2((w^Tw - 1)]\right\} \quad (7)$$
$$= \frac{1}{2} E_x[(w^Tx)^2/w + \lambda/2((w^Tw - 1)]$$

With respect to the weight vector, its gradient has the expression,

$$\frac{dj}{dw} = E_x[yx/w] + \lambda w$$

Thus the optimal multiplier may be found by vanishing $w^T \frac{dj}{dw}$, that is by solving,

$$\frac{dj}{dw} = E_x[yx/w] + \lambda ww^T = 0$$

Now the main point is to recognize that from an optimization point of view the above system is equivalent to:

$$\frac{dj}{dw} = E_x[yx/w] + \lambda\beta(ww^T - 1) = 0 , ww^T = 1$$

Where $\beta > 0$, is a constant. It can be proven that the first minor converges to the expected solution providing that the constant $\beta$ is properly chosen. This is the way to compute the optimal multiplier to obtain the stabilized learning rule [16]. The most exploited solution to the aforementioned problems consists of invoking the discrete–time versions of first minor, as

$$\Delta w = -\eta[yx - y^2 w] - \eta\beta(ww^T - 1)w , w(0) = w_0 \quad (8)$$

Where $\eta$, is the learning rate and it's a common practice to make $\eta$ a sufficiently small value which ensures good convergence in a reasonably short time which represents the discrete time stochastic counterpart of first minor rules. Neural networks MCA learning algorithms can be used to adaptively update the weight vector and reach convergence to minor component of input data. In the first order the linear MCA will be:

$$w_i(n+1) = w_i(n) - \eta y(n)[x_i(n) + y(n)w_i(n)] \quad (9)$$

For a multiple output (neuron) the output $y_j(n)$ of neuron $j$, is produced in response to the set of input,

$$x_i(n) , i = 0,1,\dots,m$$

And is given by,

$$y_j(n) = \sum_{i=1}^{m} w_{ji}(n)x_i(n) \quad (10)$$

The synaptic weight $w_{ji}$ is adapted in accordance with the generalized form of Hebbian, where the target of MCA is to extract the minor component from the input data by updating the weight vector $w(n)$ adaptively, for all $w(n) \neq 0$, as,

$$\Delta w_{ji}(n) = -\eta\left[y_j(n) x_i(n) + y_j(n) \sum_{\kappa=1}^{j} w_{ki}(n)y_k(n)\right] \quad (11)$$

Where $\Delta w_{ji}(n)$, is the change applied to the synaptic weight $w_{ji}(n)$ at time, and Examining *Eq.11*, the term, $\eta y_j(n)x_i(n)$ on the right-hand side of the equation is related to Hebbian learning. As for the second term,

$$\eta y_j(n) \sum_{k=1}^{j} w_{ki}(n)y_k(n)$$

Is related to a competitive process that goes on among the synapses in the network. Simply put, as a result of this process, the most vigorously growing (i.e., fittest) synapses or neurons are selected at the expenses of the weaker ones. Indeed, it is this competitive process that alleviates the exponential growth in Hebbian learning working by itself. Note that stabilization of the algorithm through competition requires the use of a minus sign on the right-hand side of *Eq.11*. The distinctive feature of this algorithm that it operates in a self-organized manner. This is an important characteristic of the algorithm that befits it for on-line learning. The generalized Hebbian Form of *Eq.11*, for a layer of neurons includes the algorithm of *Eq.9*, as

$$w_{ji}(n+1) = w_{ji}(n) - \Delta w_{ji} \quad (12)$$

Hence that,

$$w_{ji}(n+1) = w_{ji}(n) - \eta[y_j(n)x_i(n) + y_j(n)w_{ji}(n)] \quad (13)$$

## III. DOA MEASURMENT MODEL

### A. DOA Model

This algorithm uses measurements made on the signal received by an array of sensors. The wave fronts received by *m* sensors array element are linear combination of incident waveforms *d* and noises. The *MCA* begin with the following model of the received input data vector which is expressed as:

$$\begin{bmatrix} X_1 \\ . \\ X_m \end{bmatrix} = C(\theta_1), \dots, C(\theta_d) \begin{bmatrix} S_1 \\ . \\ S_d \end{bmatrix} + \begin{bmatrix} N_1 \\ . \\ N_m \end{bmatrix} \quad (14)$$

Where *S*, is the vector of incident signals, *N* is the noise vector and $C(\theta_d)$ is the array steering vector corresponding to the DOA of the *i*'th signal. The received vector *X* and the steering vector $C(\theta_d)$ as vector in *m* dimensional space, the input covariance matrix $R_{xx}$ can be expressed [21]:

$$R_{xx} = E[XX^T] = E[SS^T] CC^T + E[NN^T] \quad (15)$$

In many practical applications, the smallest eigenvalue of the matrix *R* of input data is usually larger than zero due to the noisy signals. The column vectors of steering vectors, is perpendicular to the eigenvector corresponding to the noise. The *MCA* spectrum may be expressed as,

$$P_{MCA}(\theta_d) = 1/[C(\theta_d)w_N w_N^T C^T(\theta_d)] \quad (16)$$

The matrix $w_N w_N^T$ is a projection matrix onto the noise subspace. For steering vectors that are orthogonal to the noise subspace, the denominator of *Eq.16*, will become very small and thus the peaks will occur in $P_{MCA}(\theta)$ corresponding to the angle of arrival of the signal. Where the ensemble average of the array input matrix *R* is known and the noise can be considered uncorrelated and identically distributed between the elements [22].

Table.1. A summary of different DOA algorithms

|   | Method | Power spectral as function of, $\theta$ |   |
|---|--------|----------------------------------------|---|
| 1 | PCA    | $C(\theta_d)R_{ss}^{-1}C(\theta_d)$    | Signal subspace |
| 2 | MCA    | $C(\theta_d)R_{NN}C(\theta_d)$         | Noise subspace |

### C. Learning Rate Parameter

The learning rate parameter has a direct effect on the convergence of the algorithm and the learning rate can have a significant effect on the accuracy.

The learning rate should be quite small $0 < \eta < 1$, otherwise the learning will become unstable and diverge. This may bring some problems [23,24], such as,

- A small learning rate gives a low learning speed.
- One should pay efforts on selecting a suitable learning rate in order to prevent learning divergence.
- The degree of fluctuation and thus the solution's accuracy will also be affected by an inappropriately predefined learning rate.

The learning rate correlated to be time-varying [25,26]. For this purpose, the learning rate usually should be set at a suitable value to reach the optimum solution of the algorithm and to move the algorithm too close in the "correct" direction.

## IV. Simulation Results

In this section we describe our simulation and results. Programs were written for DOA estimation in Matlab. A general test example is used for this purpose two sources, signal located at the far field at $(60^0, 100^0)$ degree with normalized frequencies of (0.35,0.36) fs respectively were used. A ULA of five snapshots (L), eight sensors and sensor spacing equaling half wave length $(\Delta = 0.5\ \lambda)$, spacing was used to collect the data.

### A. Effect of varying the learning rate parameter

In this simulation, we show the effect of varying the learning rate parameter has a direct effect on the convergence of the weight vector. When the learning rate has a large step size that is shown in Fig.2, it allows the algorithm to update quickly, and may also cause the estimate of the optimum solution to wander significantly until the algorithm reaches convergence and the error reaches zero. When learning rate has a small step size that is shown the convergence will be painfully slow typically. Therefore, it should be selected a suitable learning rate in order to prevent learning divergence, because this unsuitable value will make the algorithm deviate drastically from the normal learning, which may result in divergence or an increased learning time.

### B. Effect of Changing the Number of Snapshots

- Figures (4,5) show the estimated *DOA* of incoming signal. It's apparent that the spectral peaks of proposed *MCA* multiple sources become better when the number of snapshots increases, as shown in Fig.5, when the number of snapshots equal to five.
- Figures (6,7) show the estimated *DOA* of two sources for incoming signals, with changing number of snapshot. It also is apparent the spectral peaks of *PCA* become sharp and the resolution increases when the number of snapshots is increased, as shown in Fig.7, when the number of snapshots equals five.

### 2. Effect of added white noise vector

Figures (8,9) show the estimated *DOA* of two sources for incoming signals in *PCA* and proposed *MCA*, respectively, in order to compare a proposed *MCA* performance with *PCA* when the input vector is affected by white noise vector. Fig.9, shows the proposed *MCA* estimate a right angles, where the spectral has better accuracy than the *PCA* spectral plotted as shown in Fig.8.

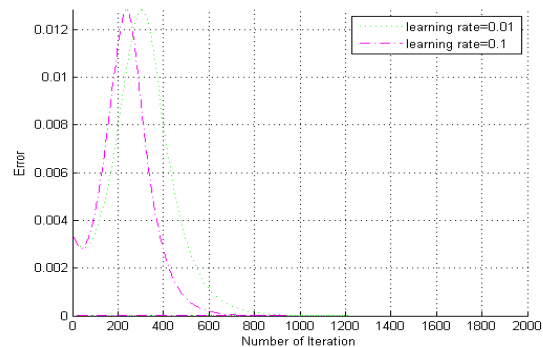

Figure.2. Learning rate step when $\eta$= (0.01 and 0.1)

### *VI. Conclusion*

This paper presented a prototype direction of arrival estimation. During this study, a MCA learning algorithm is presented to extract minor component from input, and the learning rate parameter ensures fast convergence of the algorithm. Clearly, this shows that the MCA has converged to the minor component of input signals.

Also, this demonstrate shows the MCA algorithm achieves to produce a correct angle for the DOA, when the input vector is affected by white noise vector better than the PCA algorithm, that fails to produce a value for the DOA above certain level of noise. The main advantage of this algorithm is it can better tolerate noises signals to extract the minimum eigenvalue.

REFRENCES

[1] Alexander I. Galushkin "Neural Networks Theory" Springer-Verlag Berlin Heidlberg, 2007, ISBN 0-387-94162-5 .

[2] Timo Honkela, Włodzisław Duch." Artificial Neural Networks and Machine Learning –ICANN 2011" 21st International Conference on Artificial Neural Networks Espoo, Finland, June 14-17, 2011 Proceedings.

[3] G. Dreyfus."Neural Networks, Methodology and Applications" Original French edition published by Eyrolles, springer, Paris,2004, ISBN 103-540-22980.

[4] Dovid Levin, Emanuel A., Sharon G. "Maximum Likelihood Estimation of Direction of Arrival using an acoustic vector-sensor", International Audio Laboratories Erlangen, Germany, 2012.

[5] Malcolm Hawkes "Acoustic Vector-Sensor Beamforming and Capon Direction Estimation "IEEE Transection on Signal Processing, Vol. 46, No. 9, SEPTEMBER 1998.

[6] K R SRINIVAS and V U REDDY "Sensitivity of TLS and Minimum Norm Methods of DOA Estimation to errors due to either Unite data or sensor gain and phase perturbations", Sadhana, Vol. 16,Part 3 November 1991,pp. 195-212. © Printed in India.

[7] Mitsuharu M., Shuji H. "Multiple Signal Classification by Aggregated Microphones" 2005, IEICE, ISSN: 0916-8508.

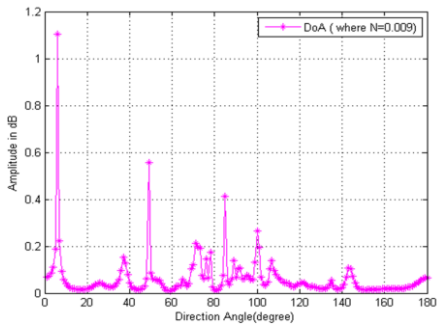
Figure (4): Estimation *DOA* by *MCA* when number of snapshots *L< 5*

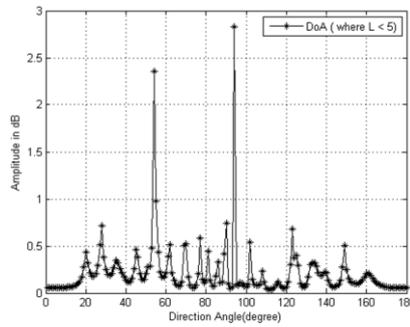
Figure (6): Estimation *DOA* by *PCA* when number of snapshots *L< 5*

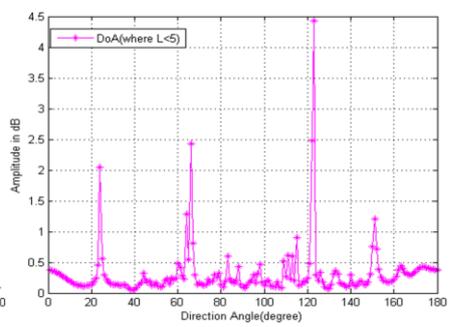
Figure (8): Estimation *DOA* by *PCA* when additive noise *N = 0.009 dB*

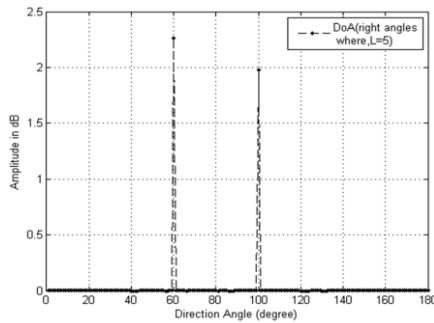
Figure (5): Estimation *DOA* by *MCA* when number of snapshots *L=5*

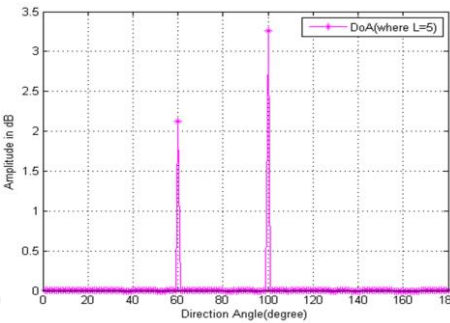
Figure (7): Estimation DOA by *PCA* when number of snapshots *L=5*

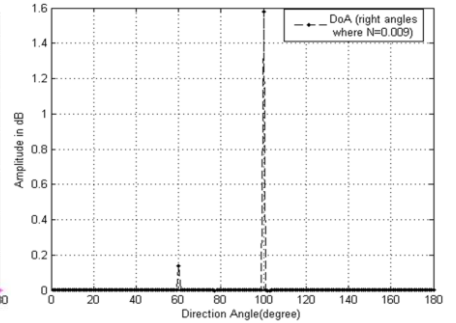
Figure (9): Estimation DOA by MCA when additive noise *N = 0.009 dB*

[8]  FeifeiGao and Alex B. Gershman "A Generalized ESPRIT Approach to Direction-of-Arrival Estimation"IEEE Signal Processing Letter, Vol. 12, No. 3, March 2005.

[9]  Pradhumna S., Michael H., Puttipong M., "Performance Analysis for Direction of Arrival Estimating Algorithms, IEEE 75th Vehicular Technology Conference(VTC Spring), 2012.

[10] G.Wang and X.-G.Xia." Iterative Algorithm for Direction of Arrival Estimation with wideband chirp signals" IEEE., 2000, ISSN :  1350-2395 .

[11] Yanwa Zhang "CGHA For Principal Component Extraction In The Complex Domain ",IEEE, Transaction on Neural Networks ,Vol. 8,no. 5,pp. 1031-1036 ,sept. 1997.

[12] Adnan S.," DOA Based Minor Component Estimation using Neural Networks", AJES, Electrical Engineering Dept., Vol.3, No.1, 2010.

[13] Kwang In Kim, Matthias O. Franz, Bernhard, "Kernel Hebbian Algorithm for Iterative Kernel Principal Component  Analysis ",Max Planck Institute for Biological Cybernetics, June 2003.

[14] K. Gao, M.O. Ahmad, M.N. Swamy, "Learning Algorithm for Total Least Squares Adaptive Signal Processing, Electronics Letters. Feb. 1992.

[15] S. Barbarossa, E. Daddio, G. Galati, "Comparison of Optimum and Linear Prediction Technique for Clutter Cancellation", Communications, Radar and Signal Processing, IEE Proceedings, ISSN (0143-7070).

[16] L. Xu, E. Oja, C. Suen, "Modified Hebbian Learning for Curve and Surface Fitting, Neural Networks, 1992.

[17] J.W. Griffiths, "Adaptive Array Processing, A Tutorial, Communications, Radar and Signal Processing, IEE Proceedings, ISSN :0143-7070.

[18] R. Schmidt, Multiple Emitter Location and Signal Parameter Estimation, IEEE Trans. Antennas Propagation (1986) 276–280.

[19] Dezhong Peng, Zhang Yi." A New Algorithm for Sequential Minor Component Analysis" International Journal of Computational Intelligence Research ,ISSN 0973-1873 Vol.2, No.2 (2006).

[20] Jie Luo, Xieting Ling "Minor Component Analysis with Independent to Blind 2 Channel Equalization," IEEE, Fudan University-China.

[21] Donghai Li, Shihai Gao, Feng Wang, "Direction of Arrival Estimation Based on Minor Component Analysis Approach", Neural Information Processing, Springer-Verlag Berlin Heidelberg , 2006.

[22] Belloni F., Richter A.,  Koivunen V. " DOA Estimation via Manifold Separation for Arbitrary Array Structures," IEEE, Transaction on signal processing, Vol, 55,No.10, October, 2007.

[23]  Qingfu Zhang, Yiu-Wung Leung, "A Class of Learning Algorithms for Principal Component Analysis and Minor Component Analysis", IEEE Transection on Neural Network, Vol. 11, No.2, March 2000.

[24] D. Randall Wilson, Tony R.," The Need for Small Learning Rates on Large Problems", International Joint Conference on Neural Networks, 2001.

[25] Tadeu N., Sergio L. Netto, Paulo S.,"Low complexity covariance-Based DOA Estimation Algorithm," EURASIP, 2007.

[26] C. Chatterjee, San Diego,"On relative convergence properties of principal component analysis algorithms", presented at IEEE Transactions on Neural Networks, 1998.